\documentclass[letterpaper]{article} 
\usepackage{aaai2026}  
\usepackage{times}  
\usepackage{helvet}  
\usepackage{courier}  
\usepackage[hyphens]{url}  
\usepackage{graphicx} 
\usepackage{natbib}  
\usepackage{caption} 
\usepackage{float}

\usepackage{amsmath}
\usepackage{amssymb}
\usepackage{amsfonts}
\usepackage{bm} 

\usepackage{multirow}

\usepackage{amssymb}
\usepackage{amsfonts}
\usepackage{bm} 
\usepackage{algpseudocode}

\usepackage{tikz}
\usetikzlibrary{shapes.geometric, arrows.meta, positioning, fit, backgrounds}

\usepackage[utf8]{inputenc}
\usepackage{graphicx}
\usepackage{booktabs} 

\usepackage{natbib}

\usepackage{algorithm}

\usepackage{newfloat}
\usepackage{listings}
\DeclareCaptionStyle{ruled}{labelfont=normalfont,labelsep=colon,strut=off} 
\floatstyle{ruled}
\newfloat{listing}{tb}{lst}{}
\floatname{listing}{Listing}
\title{MARS: A Meta-Adaptive Reinforcement Learning Framework for Risk-Aware Multi-Agent Portfolio Management}
\author{
    Jiayi Chen,
    Jing Li, 
    Guiling Wang
}
\affiliations{
    \textsuperscript{\rm }Department of Computer Science, New Jersey Institute of Technology\\
    \{jc2693, jingli, gwang\}@njit.edu
}

\begin{document}

\maketitle

\begin{abstract}

Reinforcement Learning (RL) has shown significant promise in automated portfolio management; however, effectively balancing risk and return remains a central challenge, as many models fail to adapt to dynamically changing market conditions. We propose Meta-controlled Agents for a Risk-aware System (MARS), a novel framework addressing this through a multi-agent, risk-aware approach. MARS replaces monolithic models with a Heterogeneous Agent Ensemble, where each agent’s unique risk profile is enforced by a Safety-Critic network to span behaviors from capital preservation to aggressive growth. A high-level Meta-Adaptive Controller (MAC) dynamically orchestrates this ensemble, shifting reliance between conservative and aggressive agents to minimize drawdown during downturns while seizing opportunities in bull markets. This two-tiered structure leverages behavioral diversity rather than explicit feature engineering to ensure a disciplined portfolio robust across market regimes. Experiments on major international indexes confirm that our framework significantly reduces maximum drawdown and volatility while maintaining competitive returns.

\end{abstract}

%

\section{Introduction}
\label{sec:introduction}

The application of Deep Reinforcement Learning (DRL) to automated portfolio management has undergone a significant evolution, transitioning from foundational concepts to sophisticated end-to-end trading systems. Early efforts focused on adapting classic DRL algorithms to financial decision-making, establishing DRL as a viable paradigm for learning sequential trading policies directly from market data~\cite{jiang2017deep, bai2024review}. The development of open-source libraries such as FinRL played a pivotal role during this period by standardizing the application of DRL algorithms to financial markets and improving reproducibility~\cite{liu2020finrl}.

Despite these advances, the direct application of generic DRL algorithms to financial markets exposed a fundamental mismatch. Financial environments are inherently noisy and exhibit pervasive \textbf{non-stationarity}~\cite{zhang2025adaptive}, where the statistical properties of financial time series change over time, violating the core Markov Decision Process (MDP) assumption of a stationary environment. As a result, models trained in one market condition, such as a low-volatility bull market, often fail catastrophically when the regime shifts, rendering previously learned patterns obsolete~\cite{zhang2025adaptive}. The second critical limitation is the \textbf{superficial treatment of risk}. In many DRL models, risk management is handled implicitly through reward shaping, such as using risk-adjusted metrics like the Sharpe Ratio as the reward signal or adding penalties for large drawdowns~\cite{reis2025role}. This approach is fundamentally reactive, treating risk as a consequence to be penalized after the fact, rather than as a factor to be proactively managed within the decision-making process, as human traders do. As a result, agents are often vulnerable to tail risks and sudden market shocks. Notably, these two challenges, non-stationarity and risk, are deeply interconnected: an agent that fails to adapt to changing market regimes cannot effectively manage risk.

In this paper, we propose \textbf{MARS} (\underline{M}eta-controlled \underline{A}gents for a \underline{R}isk-aware \underline{S}ystem), a novel framework that explicitly addresses the dual challenges of non-stationarity and risk. Different from the conventional monolithic agent paradigm, MARS employs a meta-learning-controlled multi-agent architecture that decouples risk preference and management from market adaptation. At the lower level, MARS employs an ensemble of heterogeneous \textbf{Safety-Critic Agents}. Each agent consists of three networks: an Actor, a Critic, and a \textbf{\textit{Safety Critic}} that learns to estimate the risk associated with a given state-action pair~\cite{srinivasan2020learning}. Crucially, each agent is explicitly configured with a distinct risk tolerance, defined by a \textbf{\textit{safety threshold}} ($\theta_i$) and a \textbf{\textit{safety weight}} ($\lambda_i$), embedding risk management directly into its learning objective in a principled and structural manner. At the higher level, a \textbf{Meta-Adaptive Controller} acts as a meta-policy, learning to dynamically orchestrate the agent ensemble. It takes the current market state as input and outputs a set of weights that determine each agent's contribution to the final trading decision. By learning to modulate these weights, the meta-controller allows the framework to adapt its aggregate strategy---ranging from ``Ultra Conservative'' to ``Maximum Growth''---in response to the non-stationary dynamics of the financial environment.
Specifically, this paper makes the following primary contributions:
\begin{enumerate}
    \item We propose \textbf{MARS}, a novel meta-learning-controlled multi-agent DRL framework featuring an ensemble of \textbf{Safety-Critic Agents} with explicit risk profiles, orchestrated by a \textbf{Meta-Adaptive Controller} to address the dual challenges of non-stationarity and risk management.
    \item We design a novel, two-part risk management mechanism specific to portfolio management: (1) a custom environmental risk score that provides the Safety-Critic a nuanced, holistic understanding of both structural and market risks, and (2) a rule-based overlay that ensures all executed actions are compliant with practical, real-world trading constraints.
    \item Extensive experiments on real-world stock market data show that MARS significantly outperforms both traditional and state-of-the-art DRL baselines, particularly in achieving higher risk-adjusted returns and preserving capital during periods of elevated market volatility.

\end{enumerate}

\section{Related Work}
\label{sec: related works}

Recent research in quantitative finance reveals a significant methodological evolution from supervised prediction to end-to-end Reinforcement Learning (RL). This shift is motivated by the ``prediction-profitability gap,'' where higher prediction accuracy does not reliably translate to better trading returns \cite{jiang2024benchmarking}, and by RL's inherent suitability for sequential decision-making. Researchers are applying increasingly sophisticated RL paradigms to tackle financial challenges like market non-stationarity and low signal-to-noise ratios \cite{liu2022finrl, wang2024quantbench}, leading to a diverse ecosystem of approaches. Concurrently, the development of standardized benchmarks like FinRL-Meta \cite{liu2022finrl} and TradeMaster \cite{sun2023trademaster} signifies a community-wide push for greater scientific rigor.

\textbf{Model-Free Approaches}. Model-free RL approaches learn trading policies directly from market interaction without an explicit market model. Recent advancements focus on augmenting standard RL agents with domain-specific architectures. For instance, DeepTrader is a risk-aware agent using a dual-module architecture to balance return with risk by embedding market conditions and penalizing high portfolio drawdown, allowing it to adapt its strategy to different market regimes \cite{wang2021deeptrader}. Addressing a different challenge, Logic-Q is a knowledge-guided system that injects human-like trading logic into a DRL agent via program sketching. This helps the agent identify major market trends and prevent catastrophic errors during trend shifts, thereby improving robustness \cite{li2025logicq}.

\textbf{Model-Based and Hybrid Approaches}. Model-based and hybrid approaches integrate predictive components to provide the RL agent with a richer understanding of the market, aiming to improve sample efficiency. A prime example is StockFormer, which fuses a predictive coding module with an RL agent, using specialized Transformer branches to learn latent representations of future dynamics for a Soft Actor-Critic (SAC) agent \cite{gao2023stockformer}. This end-to-end system tackles the low signal-to-noise problem by extracting predictive signals, though its performance can be contingent on the accuracy of the underlying predictive model. Other hybrid methods, like ``Ambiguous'' Mean-Variance RL, fuse RL with classical financial theory, using RL to learn unknown statistical parameters required by traditional risk models \cite{huang2020twolevel}.

\textbf{Hierarchical and Multi-Agent RL Approaches}. Hierarchical and Multi-Agent RL approaches decompose complex financial problems into more manageable sub-tasks. Hierarchical Reinforcement Learning (HRL) is particularly effective for multi-scale decision-making. For example, HRPM uses a two-level hierarchy where a high-level agent sets strategic portfolio allocations and a low-level agent minimizes trade execution costs, directly addressing frictions like slippage \cite{wang2021commission}. EarnHFT applies a more complex three-tier hierarchy to high-frequency trading, using a meta-controller to dynamically select the best-specialized agent for current market conditions \cite{qin2024earnhft}. Separately, Multi-Agent RL (MARL) models strategic interactions. The MAPS framework, for instance, uses cooperative agents with a diversification penalty to encourage varied strategies, creating a more robust ``portfolio of portfolios'' \cite{lee2020maps}. These approaches acknowledge that a single agent may be insufficient to capture multifaceted market dynamics.

\section{Methodology}
\label{sec:methodology}

\begin{figure*}[t]
    \centering
    
    \includegraphics[width=0.95\textwidth]{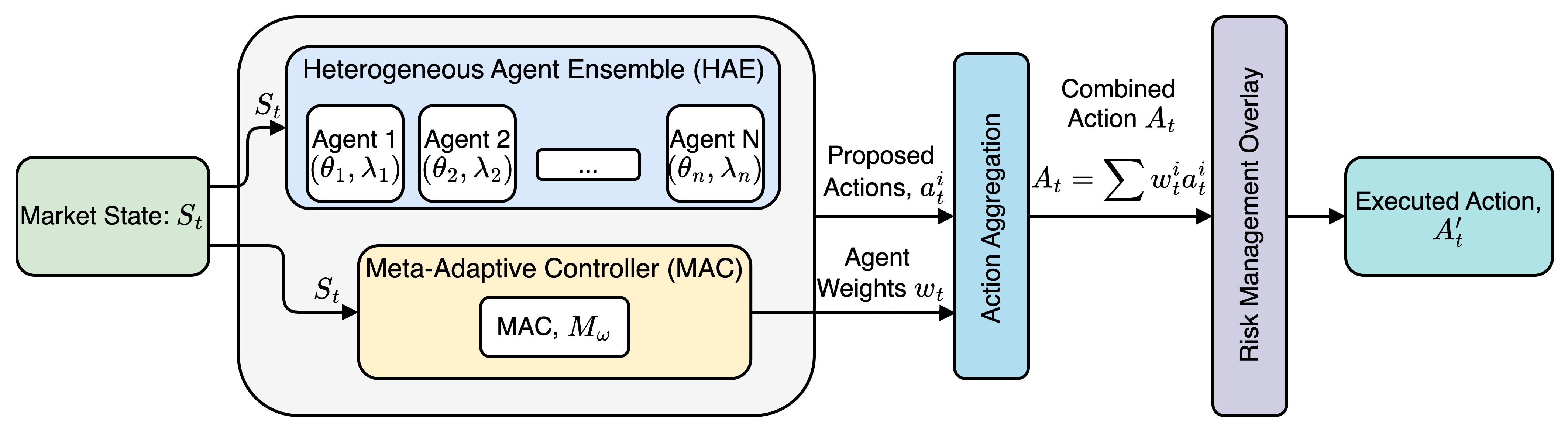}
    
    \caption{The MARS framework architecture. The system processes the Market State ($s_t$) through two parallel components. The Meta-Adaptive Controller (MAC) produces agent weights ($w_t$), while the Heterogeneous Agent Ensemble (HAE) generates proposed actions ($a_t^i$). These outputs are aggregated and passed through a Risk Management Overlay to produce the final executed action ($A'_t$).}
    \label{fig:mars_framework}
    
\end{figure*}

MARS tackles the challenges of non-stationarity and risk in financial markets by learning to adapt to the market conditions through an ensemble of RL agents with diverse risk profiles. As illustrated in Figure~\ref{fig:mars_framework}, the MARS framework takes as input the Market State vector $s_t$ at time $t$, comprising the current portfolio holdings, cash balance, and a set of technical indicators for all assets.
This state vector is concurrently fed into two main components: the Meta-Adaptive Controller (MAC) and the Heterogeneous Agent Ensemble (HAE). In each time step $t$, the MAC processes $s_t$ to generate a vector of agent weights, $w_t$, tailoring the influence of each agent to the current market condition. Simultaneously, the HAE, which consists of $N$ distinct Safety-Critic agents with unique risk profiles ($\theta_i, \lambda_i$ for agent $i$), maps $s_t$ into a set of diverse proposed actions $\{a_t^i\}_{i=1}^N$. Each agent in the ensemble is a complete DDPG-based agent composed of Actor, Critic, and Safety-Critic networks. The outputs from these two components---the agent weights and the proposed actions---are then combined via a weighted sum ($A_t$), which is passed through a final Risk Management Overlay to produce the executed action ($A'_t$).

\subsection{Problem Formulation}

We formulate the portfolio management task as a Markov Decision Process (MDP), defined by the following tuple $\mathcal{M} = (\mathcal{S}, \mathcal{A}, \mathcal{P}, \mathcal{R}, \gamma)$.

\textbf{State Space $\mathcal{S}$}: A state $s_t \in \mathcal{S}$ at time step $t$ is a comprehensive, flattened vector representation of the market environment and the agent's portfolio. It is constructed by concatenating the current cash balance $b_t$, and for each of the $D$ assets, the current holdings $h_t^i$ and a feature vector $\mathbf{x}_t^i \in \mathbb{R}^{K}$ of market data. The feature vector for each asset contains its price and 4 technical indicators (MACD, RSI, CCI, ADX). 

    \textbf{Action Space $\mathcal{A}$}: The action $A_t \in [-1, 1]^D$ is a continuous vector where each element represents the target change in allocation for one of the assets. This normalized output is then scaled by a maximum trade size to determine the actual number of shares to trade. The final executed trades are subject to risk management constraints, as detailed below in Section: Trading Procedure. 

    \textbf{Reward Function $\mathcal{R}$}: The reward $R_t$ at each step is designed to promote profit generation while penalizing risk and transaction friction. It is defined as:
    $$R_t = \frac{V_{t+1} - V_t}{V_t} - C_t - \rho_t$$
    where $\frac{V_{t+1} - V_t}{V_t}$ is the portfolio's rate of return from $t$ to $t+1$, $C_t$ is the total monetary transaction cost incurred at the step, and $\rho_t$ is a risk-aversion penalty based on the portfolio's recent performance, calculated as:
    $$\rho_t = w_{vol} \cdot \sigma_{30d} + w_{dd} \cdot DD_{30d}$$
    where $\sigma_{30d}$ and $DD_{30d}$ are the rolling 30-day portfolio volatility and max drawdown.


The objective is to learn a policy $\pi(A_t|s_t)$ that maximizes the expected cumulative discounted reward:
$$J(\pi) = \mathbb{E}_{\tau \sim \pi} \left[ \sum_{t=0}^{T} \gamma^t R_t \right]$$
where 
$\gamma$ 
is the discount factor.

\subsection{Overall Training Algorithm}

The complete training and evaluation procedure for the MARS framework is summarized in Algorithm \ref{alg:mars}.

\begin{algorithm}[!ht]
\caption{MARS Training and Evaluation Algorithm}
\label{alg:mars}
\begin{algorithmic}[1]
\State Initialize HAE agents $\{\mathcal{A}_i\}_{i=1}^N$ with networks $\pi_{\phi_i}, Q_{\psi_i}, C_{\xi_i}$
\State Initialize MAC controller $M_{\omega}$
\State Initialize replay buffers $\mathcal{D}_i$ for each agent and $\mathcal{D}_M$ for the MAC
\For{episode = 1 to max\_episodes}
    \State Reset portfolio and environment to get initial state $s_0$
    \For{$t = 0$ to $T-1$}
        \State \Comment{Decision Making}
        \State Get individual actions $a_t^i = \pi_{\phi_i}(s_t)$ from each agent $\mathcal{A}_i$
        \State Get agent weights $\mathbf{w}_t = \text{softmax}(M_{\omega}(s_t))$ from MAC
        \State Aggregate final action $A_t = \sum_{i=1}^{N} w_t^i \cdot a_t^i$
        \State Apply risk management overlay to get final trade $A'_t = \text{validate}(A_t, s_t)$
        \State \Comment{Environment Interaction}
        \State Execute $A'_t$, observe reward $R_t$ and next state $s_{t+1}$
        \State Store transition $(s_t, A'_t, R_t, s_{t+1})$ in each agent's replay buffer $\mathcal{D}_i$
        \State Store $(s_t, \{Q_{\psi_i}(s_t, a_t^i)\}_{i=1}^N, \{C_{\xi_i}(s_t, a_t^i)\}_{i=1}^N)$ in MAC buffer $\mathcal{D}_M$
        \State \Comment{Agent Training}
        \State For each agent $\mathcal{A}_i$, sample a minibatch from $\mathcal{D}_i$
        \State Update critic $Q_{\psi_i}$, safety-critic $C_{\xi_i}$, and actor $\pi_{\phi_i}$
        \State Update target networks for each agent
    \EndFor
    \State \Comment{Meta-Controller Training}
    \If{episode mod meta\_train\_freq == 0}
        \State Sample a minibatch from $\mathcal{D}_M$
        \State Update MAC controller $M_{\omega}$ by minimizing $\mathcal{L}(\omega)$
    \EndIf
\EndFor
\end{algorithmic}
\end{algorithm}

\subsection{Heterogeneous Agent Ensemble (HAE)}

The core of our framework is an ensemble $\mathcal{E} = \{ \mathcal{A}_1, \mathcal{A}_2, ..., \mathcal{A}_N \}$ of $N$ distinct agents. Each agent $\mathcal{A}_i$ is a complete Safety-Critic agent defined by a unique, intrinsic risk profile $(\theta_i, \lambda_i)$, where $\theta_i$ is its risk tolerance threshold and $\lambda_i$ is its risk aversion penalty. This heterogeneity is a key design choice, creating a diverse pool of ``expert" behaviors ranging from ultra-conservative to highly aggressive.

Each agent $\mathcal{A}_i$ is implemented using a Deep Deterministic Policy Gradient (DDPG) architecture, extended with a dedicated Safety-Critic network. 

\subsubsection{Actor Network $\pi_{\phi_i}(s_t)$}
The actor, a Multi-Layer Perceptron (MLP) with parameters $\phi_i$, maps the state $s_t$ to a deterministic action $a^i_t$. It is updated via a policy gradient that includes a novel \textbf{Conditional Safety Penalty (CSP)}:
\begin{equation*}
\begin{split}
\nabla_{\phi_i}J(\phi_i) & \approx \mathbb{E}_{s_t \sim \mathcal{D}} \Big[ \nabla_{\phi_i} Q_{\psi_i}(s_t, \pi_{\phi_i}(s_t)) \\
& \qquad - \lambda_i \cdot \nabla_{\phi_i} \text{ReLU}\big(C_{\xi_i}(s_t, \pi_{\phi_i}(s_t)) - \theta_i\big) \Big]
\end{split}
\end{equation*}
The CSP term explicitly penalizes the policy only when its proposed action's predicted risk $C_{\xi_i}$ exceeds its specific risk tolerance $\theta_i$.

\subsubsection{Critic Network $Q_{\psi_i}(s_t, a_t)$}
The critic, an MLP with parameters $\psi_i$, approximates the state-action value function by training to minimize the Temporal Difference (TD) error:
$$ \mathcal{L}(\psi_i) = \mathbb{E}_{(s_t, a_t, R_t, s_{t+1}) \sim \mathcal{D}} \left[ \left( y_t - Q_{\psi_i}(s_t, a_t) \right)^2 \right] $$
where the TD target $y_t = R_t + \gamma Q_{\psi'_i}(s_{t+1}, \pi_{\phi'_i}(s_{t+1}))$.

\subsubsection{Safety-Critic Network $C_{\xi_i}(s_t, a_t)$}
This network, with parameters $\xi_i$, is architecturally similar to the critic but is trained to predict the extrinsic risk of an action. Its objective is to learn an environment risk function, $\mathcal{C}_{env}$. This target function, a novel component of our framework, is specifically designed to measure risk in a stock trading context by computing a score in $[0, 1]$ based on three key metrics: \textbf{Portfolio Concentration} (penalizing over-concentration), \textbf{Leverage} (quantifying reliance on borrowed funds), and \textbf{Simulated Volatility} (estimating a trade's forward-looking risk impact). 
Based on sensitivity analysis, the final score uses fixed weights of 40\% Simulated Volatility, 30\% Portfolio Concentration, and 30\% Leverage, which were tuned to achieve a balance across different market regimes.
By integrating these distinct risk dimensions, $\mathcal{C}_{env}$ provides the Safety-Critic with a holistic and financially-grounded risk signal that goes beyond simple price-based penalties. The Safety-Critic is trained via a Mean Squared Error loss against this target:
$$ \mathcal{L}(\xi_i) = \mathbb{E}_{(s_t, a_t) \sim \mathcal{D}} \left[ \left( \mathcal{C}_{env}(s_t, a_t) - C_{\xi_i}(s_t, a_t) \right)^2 \right] $$

\subsection{Meta-Adaptive Controller (MAC)}

The Meta-Adaptive Controller, $M_{\omega}$, serves as a high-level orchestrator. It is a neural network with parameters $\omega$ that learns a meta-policy, $\pi_\omega(\mathbf{w}_t | s_t)$, which dynamically assigns weights to the agents in the HAE based on the current market state $s_t$. This allows the framework to implicitly learn and adapt to different market regimes (e.g., bull, bear, volatile) by adjusting its reliance on different risk-taking behaviors.

The controller outputs a vector of logits, passed through a softmax function to generate the weight distribution:
$$ \mathbf{w}_t = [w_t^1, ..., w_t^N] = \text{softmax}(M_{\omega}(s_t)) $$

The final action $A_t$ is an aggregation of the individual agents' proposed actions, weighted by MAC's output:
$$ A_t = \sum_{i=1}^{N} w_t^i \cdot \pi_{\phi_i}(s_t) $$

The MAC is trained to maximize a risk-adjusted utility function. The loss is the negative of a custom objective that balances the mean and standard deviation of the ensemble's predicted Q-values (a Sharpe-like term) while also penalizing the ensemble's predicted risk:
$$ \mathcal{L}(\omega) = - \left( \frac{\mathbb{E}[\bar{Q}_t]}{\text{Std}(\bar{Q}_t) + \epsilon} - \lambda_{meta} \cdot \mathbb{E}[\bar{C}_t] \right) $$
where $\bar{Q}_t = \sum_{i=1}^{N} w_t^i Q_{\psi_i}(s_t, a_t^i)$ is the weighted-average predicted Q-value, $\bar{C}_t = \sum_{i=1}^{N} w_t^i C_{\xi_i}(s_t, a_t^i)$ is the weighted-average predicted risk, and $\lambda_{meta}$ is a hyperparameter. 
By minimizing this objective, the MAC learns to favor agent combinations that promise high, stable returns with low predicted risk, effectively navigating the risk-return tradeoff for the entire system.

\subsection{Trading Procedure}

The decision-making and trading process at each time step $t$ begins with the construction of the state vector $s_t$ from the latest market data. It is important to note that the Safety-Critic network is a \textbf{training-only module} used to instill each agent's intrinsic risk profile; during deployment (i.e., this trading procedure), risk is managed by the MAC's dynamic weighting and the final rule-based overlay. Concurrently, each agent $\mathcal{A}_i$ in the HAE proposes an individual action $a^i_t$, while the Meta-Adaptive Controller $M_{\omega}$ generates the corresponding agent weight vector $\mathbf{w}_t$. These outputs are then combined via action aggregation, where the final system action $A_t$ is computed as the weighted average of all proposed actions. Before execution, $A_t$ is passed through a risk management overlay. This overlay acts as a final failsafe to ensure all actions are practical and compliant with institutional standards by enforcing rules such as limits on position concentration, maintenance of a cash buffer for liquidity, and a ban on short-selling. This rule-based system provides hard guardrails against diversification risk and unlimited losses, bridging the gap between the agent's learned policy and real-world trading compliance. Any action violating these rules is adjusted to produce a final, risk-compliant action, $A'_t$. Finally, this action is executed in the market, and the environment transitions to the next state $s_{t+1}$.

\begin{table*}[tb]
\centering
\small
\begin{tabular}{@{}lllll@{}}
\toprule
\textbf{Test Period} & \textbf{Training Span} & \textbf{Validation Span} & \textbf{Testing Span} & \textbf{Primary Market Condition} \\ \midrule
2022 Test & 2016-01-01 to 2020-12-31 & 2021-01-01 to 2021-12-31 & 2022-01-01 to 2022-12-31 & Volatile Bear Market \\
2024 Test & 2018-01-01 to 2022-12-31 & 2023-01-01 to 2023-12-31 & 2024-01-01 to 2024-12-31 & Recent Bull Market  \\ \bottomrule
\end{tabular}
\caption{Training and Testing Periods for Datasets}
\label{tab:datasets}
\end{table*}

\begin{table}[tb]
\centering  
\footnotesize
\setlength{\tabcolsep}{3.4pt} 

\begin{tabular}{llccccc}
\toprule
\textbf{Env.} & \textbf{Model} & \textbf{CR\%} & \textbf{AR\%} & \textbf{SR} & \textbf{AVol\%} & \textbf{MDD\%} \\ 
\midrule
\multirow{9}{*}{\textbf{\shortstack[l]{DJI\\2024}}}
 & \textbf{MARS} & \textbf{29.50} & \textbf{31.19} & \textbf{2.84} & 10.99 & \textbf{-5.39} \\
 & MARS-Static & 17.10 & 17.17 & 1.71 & \textbf{10.04} & -6.79 \\
 & MARS-Homo & 22.21 & 22.31 & 1.85 & 12.03 & -7.81 \\
 & MARS-Div5 & 12.02 & 12.07 & 1.08 & 11.16 & -6.19 \\
 & MARS-Div15 & 19.70 & 19.87 & 1.67 & 11.89 & -7.26 \\
 & DeepTrader & 13.30 & 14.01 & 1.18 & 11.92 & -6.84 \\
 & HRPM & 19.11 & 20.16 & 0.99 & 20.43 & -7.90 \\
 & AlphaStock & 22.13 & 23.36 & 1.18 & 19.78 & -10.24 \\
 & DJI Index & 15.36 & 16.19 & 1.41 & 11.51 & -6.06 \\
\midrule
\multirow{5}{*}{\textbf{\shortstack[l]{HSI\\2024}}}
 & \textbf{MARS} & 16.24 & 17.84 & \textbf{1.49} & 12.00 & -7.38 \\
 & DeepTrader & 13.35 & 14.65 & 0.64 & 23.04 & -15.44 \\
 & HRPM & 4.68 & 5.12 & 0.80 & 6.38 & -6.00 \\
 & AlphaStock & -0.19 & -0.21 & -0.05 & \textbf{4.44} & \textbf{-4.63} \\
 & HSI Index & \textbf{24.46} & \textbf{28.78} & 1.10 & 26.08 & -17.09 \\
\midrule
\multirow{5}{*}{\textbf{\shortstack[l]{DJI\\2022}}}
 & \textbf{MARS} & \textbf{-0.86} & \textbf{-0.93} & \textbf{-0.05} & \textbf{19.83} & \textbf{-16.77} \\
 & DeepTrader & -10.70 & -11.43 & -0.46 & 25.07 & -21.32 \\
 & HRPM & -3.34 & -3.58 & -0.18 & 20.35 & -17.30 \\
 & AlphaStock & -36.37 & -38.42 & -1.03 & 37.35 & -46.17 \\
 & DJI Index & -3.14 & -3.36 & -0.17 & 20.26 & -19.69 \\
\midrule
\multirow{5}{*}{\textbf{\shortstack[l]{HSI\\2022}}}
 & \textbf{MARS} & \textbf{-14.50} & \textbf{-14.88} & -0.66 & \textbf{22.56} & \textbf{-32.72} \\
 & DeepTrader & -26.69 & -27.34 & -0.86 & 31.93 & -48.02 \\
 & HRPM & -18.98 & -19.46 & -0.77 & 25.21 & -37.01 \\
 & AlphaStock & -24.32 & -25.01 & \textbf{-0.64} & 39.32 & -54.60 \\
 & HSI Index & -19.77 & -21.36 & \textbf{-0.64} & 33.32 & -41.07 \\
\bottomrule
\end{tabular}
\caption{Risk–return metrics for MARS and baselines across all environments. Best results in bold.}
\label{tab:main_results}
\end{table}

\section{Experiments}

To rigorously evaluate the MARS framework, we design a comprehensive set of experiments aimed at answering the following key research questions:

\begin{enumerate}
    \item How does MARS perform against traditional passive strategies and state-of-the-art reinforcement learning agents, especially in terms of risk-adjusted returns across varying market conditions?
    \item To what extent are the core architectural components---the Heterogeneous Agent Ensemble (HAE) and the Meta-Adaptive Controller (MAC)---necessary for achieving the observed performance?
    \item Can MARS effectively and adaptively respond to market fluctuations to achieve both high returns and robust risk management?
\end{enumerate}

\subsection{Experimental Setup}

\subsubsection{Datasets}
We used historical daily data from two major global indices, the \textbf{Dow Jones Industrial Average (DJI)} and the \textbf{Hang Seng Index (HSI)}, sourced from Yahoo Finance. For the main MARS experiment and baseline comparisons, we selected a portfolio of 50 representative US stocks that benchmarked against the DJI and 50 constituent stocks from the HSI. To ensure robust evaluation across diverse market regimes, we defined two distinct time periods for training and testing, as detailed in Table~\ref{tab:datasets}. These periods were chosen to represent a volatile bear market (2022) and a more recent bull market (2024).

\subsubsection{Evaluation Metrics}
Following standard practice, we assess performance using metrics from three categories:
\begin{itemize}
    \item \textbf{Profit Criterion:} Cumulative Return (CR) and Annualized Return (AR).
    \item \textbf{Risk Criterion:} Annualized Volatility (AVol) and Maximum Drawdown (MDD).
    \item \textbf{Risk-Adjusted Return:} Sharpe Ratio (SR).
\end{itemize}

\subsubsection{Baseline Methods}
We compare MARS against a passive investment strategy and three state-of-the-art DRL models:
\begin{itemize}
    \item \textbf{Market Index:} A buy-and-hold strategy for the respective index.
    \item \textbf{DeepTrader:} A DRL model that incorporates market conditions \cite{wang2021deeptrader}.
    \item \textbf{HRPM:} A hierarchical RL framework for portfolio allocation \cite{wang2021commission}.
    \item \textbf{AlphaStock:} An investment strategy using an interpretable attention network \cite{wang2019alphastock}.
\end{itemize}

\subsubsection{MARS Variants for Ablation Study}
To isolate the contributions of key components, we evaluate several variants of our MARS framework:
\begin{itemize}
    \item \textbf{MARS-Static:} The MAC is replaced with fixed, uniform agent weights.
    \item \textbf{MARS-Homogeneous:} The HAE is replaced with an ensemble of agents sharing identical risk profiles, but each is initialized with a different random seed to ensure network weight diversity.
    \item \textbf{MARS-Divergence (5/15):} The number of agents is changed to 5 or 15 to test ensemble size sensitivity.
\end{itemize}

\subsubsection{Implementation Details}
Our MARS framework was implemented using the following configuration. Hyperparameters were tuned based on performance on the validation set. For all experiments, we used a fixed random seed of 42 to ensure the reproducibility of our results. The Heterogeneous Agent Ensemble (HAE) consists of $N=10$ agents, whose risk profiles were defined by a spectrum of safety thresholds ($\theta_{i}$) ranging from 0.10 (conservative) to 0.55 (aggressive) and penalty weights ($\lambda_{i}$) from 1.0 to 5.5. For each market index, we used a portfolio of $D=50$ assets. The feature vector for each asset includes its price and 4 technical indicators (MACD, RSI, CCI, ADX).

For the reward function, the risk-aversion penalty weights were set to $w_{vol}=0.5$ and $w_{dd}=2.0$. The discount factor $\gamma$ was set to 0.99. In the Meta-Adaptive Controller's loss function, the risk penalization hyperparameter $\lambda_{meta}$ was set to 0.5. 
All networks for the actors, critics, Safety-Critics, and the MAC were implemented as fully-connected Multi-Layer Perceptrons (MLPs). The network architecture consists of three hidden layers with dimensions of 256, 128, and 64, with ReLU as the activation function applied after each hidden layer.
For the trading procedure, the position concentration limit was set to 20\% of the total portfolio value.

\begin{figure}[tb]
    \centering
    \includegraphics[width=\columnwidth]{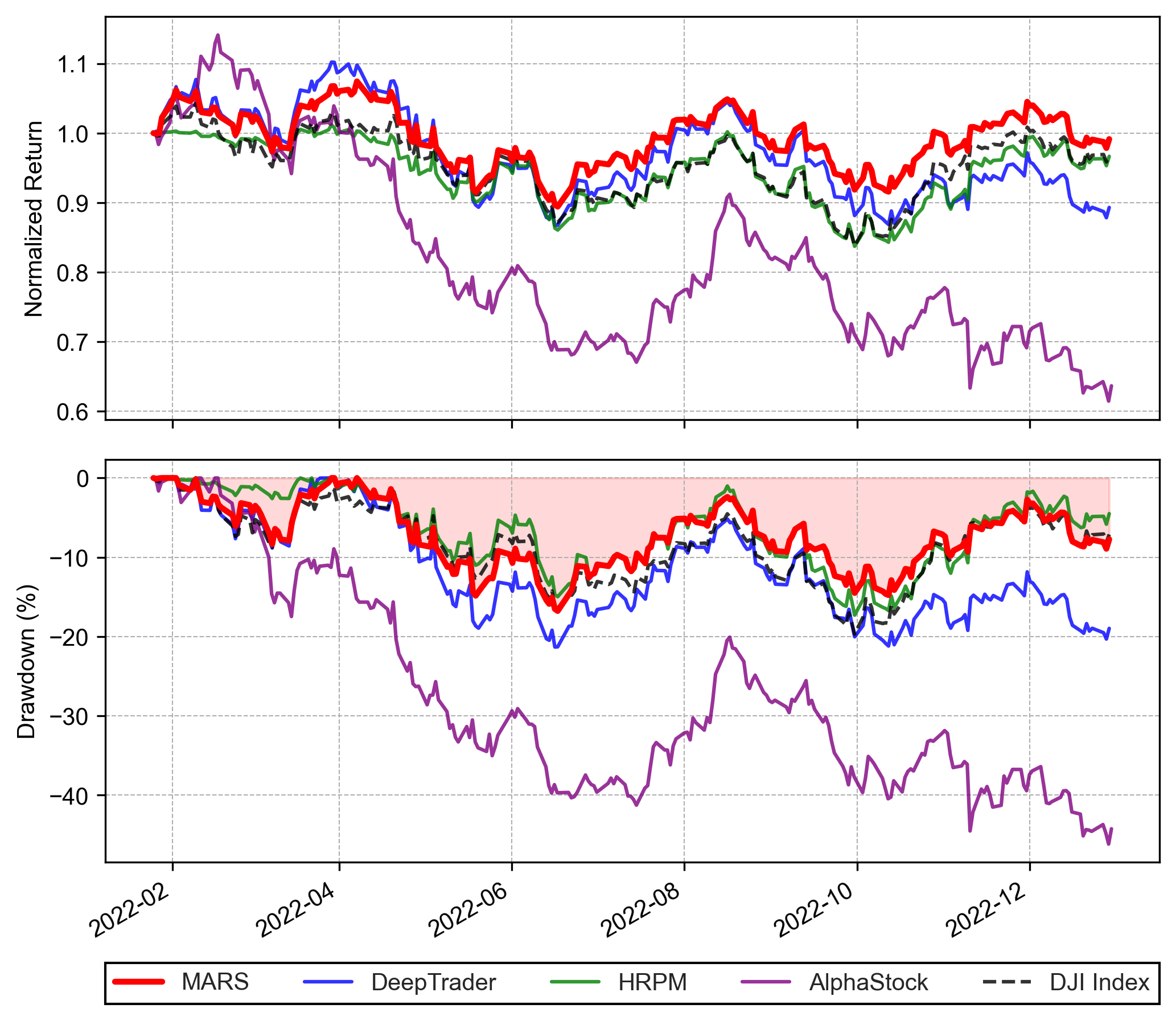}
    \caption{Performance comparison on the DJI 2022 dataset. MARS (red) shows superior capital preservation with a significantly shallower drawdown compared to baselines.}
    \label{fig:dji_2022_performance}
\end{figure}

\begin{figure}[tb]
    \centering
    \includegraphics[width=\columnwidth]{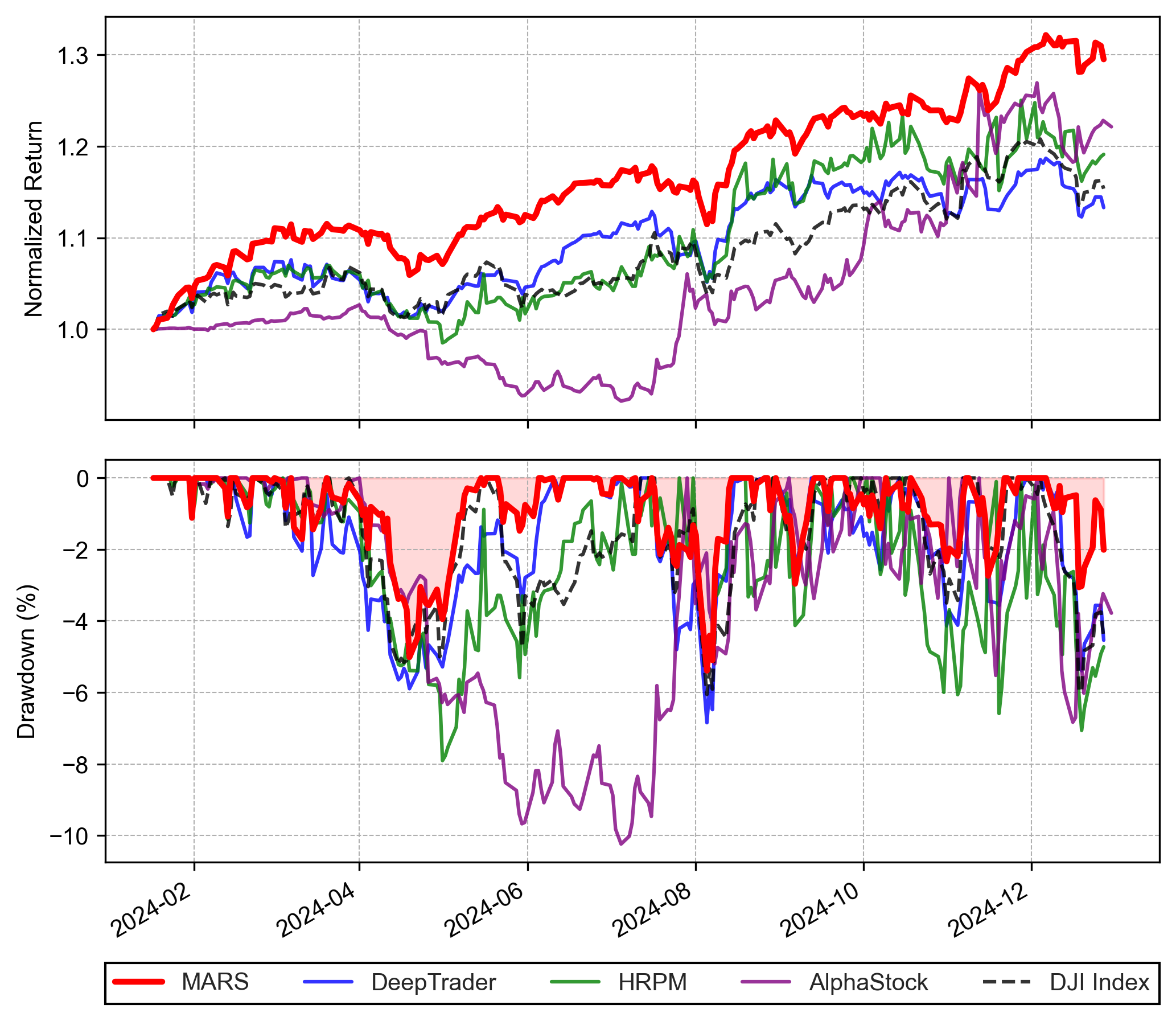}
    \caption{Performance comparison on the DJI 2024 dataset. MARS (red) achieves the highest return while maintaining a competitive drawdown profile.}
    \label{fig:dji_2024_performance}
\end{figure}
\subsection{Overall Performance}

Table~\ref{tab:main_results} summarizes the performance of MARS compared to baseline models across diverse markets.

\subsubsection{Performance on DJIA}
In the Dow Jones Industrial Average (DJIA) environments, MARS consistently delivered strong performance. During the challenging 2022 bear market, it demonstrated superior performance across all metrics, achieving the lowest loss (CR -0.86\%) and the best maximum drawdown (-16.77\%). During the more favorable 2024 bull market, MARS maintained its dominance, achieving the highest Cumulative Return (29.50\%), Annual Return (31.19\%), Sharpe Ratio (2.84), and the lowest Maximum Drawdown (-5.39\%). Notably, compared to the best baseline results, MARS achieved relative improvements of 70.6\% and 101.4\% in Sharpe Ratio for DJI 2022 and 2024, respectively.

\subsubsection{Performance on HSI}

In the 2022 bear market for the Hang Seng Index (HSI), MARS again excelled in capital preservation, securing the best Cumulative Return (-14.50\%), lowest volatility (22.56\%), and smallest Maximum Drawdown (-32.72\%). It also achieves comparable performance in Sharpe Ratio (-0.66) to the best baselines (-0.64). In the 2024 bull market, although the passive HSI Index yielded a higher raw return, MARS outperformed all DRL-based methods in both cumulative and annual returns. Moreover, MARS achieved the highest Sharpe Ratio across all models---a 35.5\% relative improvement---demonstrating the more effective balance between risk and reward.

\subsubsection{Performance during Bear Market (2022) vs. Bull Market (2024)}
Figures~\ref{fig:dji_2022_performance} and~\ref{fig:dji_2024_performance} illustrate the returns and drawdowns of different methods on the DJI during 2022 and 2024. Unlike models that follow a uniform strategy, MARS adapts its behavior to shifting market dynamics. For instance, during the volatile declines from March to June 2022 and again between August and October 2022, MARS prioritized capital preservation. This defensive posture enabled it to withstand turbulence without suffering the deep drawdowns and sharp losses seen in models such as AlphaStock and DeepTrader. By successfully mitigating the year’s two major downturns, MARS closed 2022 with the smallest overall loss.
Even in the positive market of 2024, MARS maintained vigilance, leveraging its risk-aware agents to protect gains against short-term volatility, like from April to May 2024. This resulted in controlled, minimal drawdowns compared to the sharper dips of HRPM and AlphaStock. As the bullish trend solidified in 2024, MARS shifted its strategy, giving more weight to its aggressive, growth-oriented agents, enabling it to capitalize on strong market momentum. As a result, its performance accelerated and began to diverge from competing models.
MARS’s ability to both defend against downturns and aggressively capture upside underpins its superior performance---delivering the highest cumulative return, a competitive drawdown profile, and ultimately the best Sharpe Ratio among all tested models.

\subsection{Ablation Study}
\begin{figure}[tb]
    \centering
    \includegraphics[width=\columnwidth]{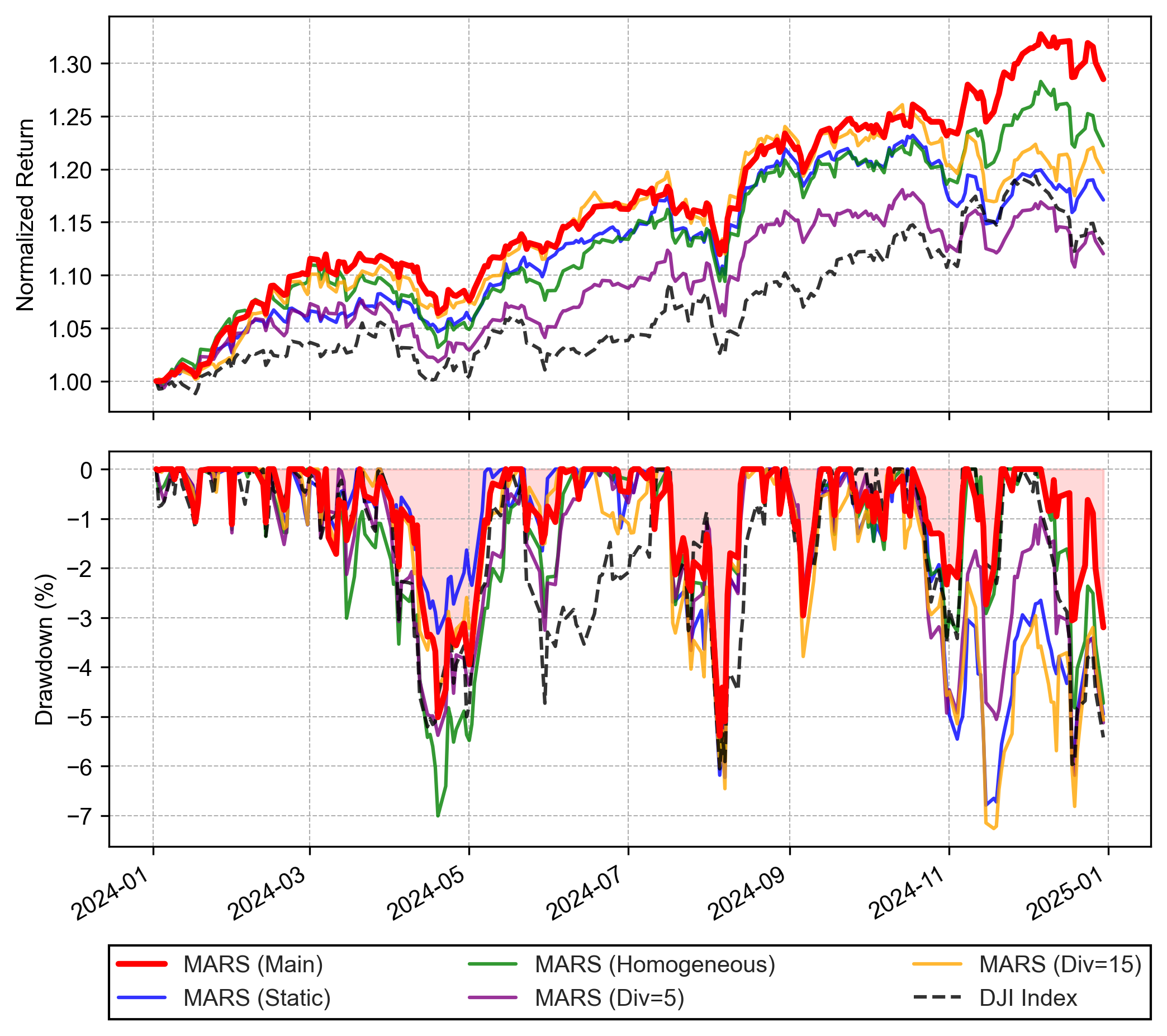}
    \caption{Ablation study performance on the DJI 2024 dataset. The main MARS model (red) outperforms all variants, validating its architectural components.}
    \label{fig:dji_ablation_performance}
\end{figure}

To assess the necessity of MARS’s core architectural components, we conducted ablation studies using DJI 2024. 
Figure~\ref{fig:dji_ablation_performance} illustrates the return and drawdown trends of the MARS variants compared with the DJI Index.

\subsubsection{Effectiveness of Meta-Adaptive Controller (MAC)}
The \texttt{MARS-Static} variant, which removes MAC’s dynamic agent weighting, performs markedly worse than the full MARS framework. Its Cumulative Return falls from 29.50\% to 17.10\%, and its Sharpe Ratio drops from 2.84 to 1.71. 
This confirms the MAC's dynamic orchestration is critical for market adaptation and maximizing risk-adjusted returns.

\subsubsection{Effectiveness of Heterogeneous Agent Ensemble (HAE)}
The \texttt{MARS-Homogeneous} variant, which removes agent heterogeneity, underperforms the full model with a cumulative return of only 22.21\% and a Sharpe Ratio of 1.85. Its maximum drawdown (-7.81\%) is also worse than both the full model and \texttt{MARS-Static}, underscoring the importance of diverse risk profiles in the HAE for effectively managing downside risk.

\subsubsection{Impact of Ensemble Diversity}
To examine the impact of ensemble size, we varied the number of agents from 10 to 5 (\texttt{MARS-Div5}) and 15 (\texttt{MARS-Div15}). With only 5 agents, the Cumulative Return dropped to 12.02\%, reflecting insufficient strategic diversity. Expanding to 15 agents improved performance to 19.70\% but still fell short of the main model, indicating diminishing returns. 

\subsection{Analysis of Adaptive Strategy}

\begin{figure}[tb]
    \centering

    \includegraphics[width=\columnwidth]
{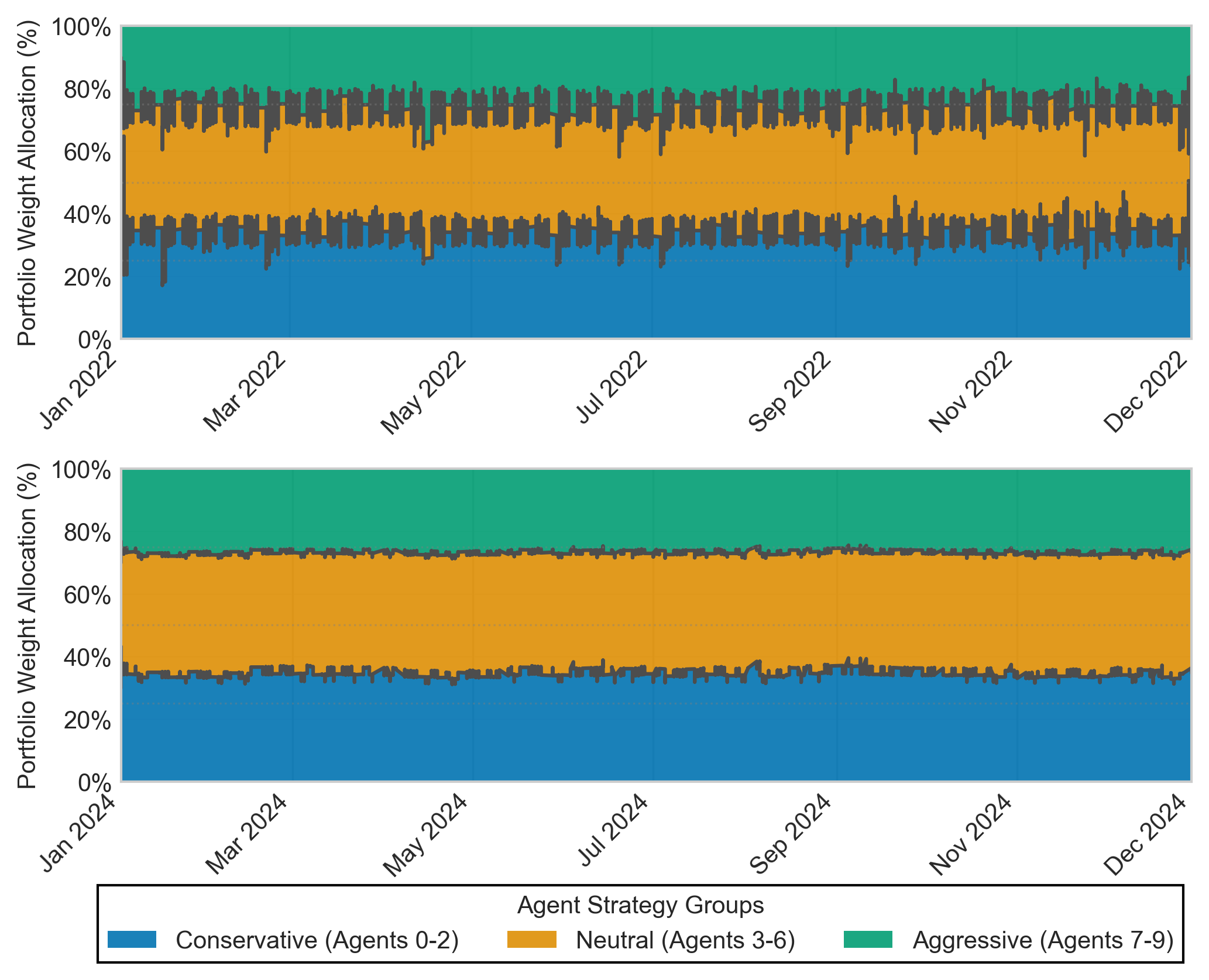}
    \caption{Comparison of agent allocation strategies under Meta-Adaptive Controller (MAC) during the 2022 bear market (top) and the 2024 bull market (bottom) for DJI portfolio.}
    \label{fig:adaptive_strategy_2024}
\end{figure}

To reveal MARS’s adaptive capabilities qualitatively, we analyzed the behavior of the Meta-Adaptive Controller (MAC) under contrasting market conditions—the volatile 2022 market and the stable 2024 market—using the DJI portfolio. Figure~\ref{fig:adaptive_strategy_2024} shows the time-varying weights that MAC assigned to the Conservative, Neutral, and Aggressive agent groups.

The results reveal two distinct meta-strategies. In the turbulent 2022 bear market (top panel), the MAC adopted a highly reactive, defensive posture, with allocation weights showing substantial day-to-day volatility. The Aggressive group’s allocation volatility was over 70\% higher than in 2024, and the MAC frequently shifted between Conservative and Neutral agents—reflecting a dynamic strategy aimed at navigating uncertainty and mitigating risk.

In contrast, during the 2024 bull market (bottom panel), the MAC settled into a remarkably stable and confident meta-strategy. Daily weight fluctuations were smoother and far less volatile, while mean allocations remained similar to 2022 (roughly 34.6\% Conservative, 38.6\% Neutral, 26.9\% Aggressive). Notably, coordination between groups strengthened: the negative correlation between Conservative and Aggressive allocations deepened from -0.788 in 2022 to -0.968 in 2024, indicating a more decisive, synchronized trade-off between risk and growth.

This comparison confirms that MAC does not employ a static policy but instead fundamentally adapts its operational behavior in response to the market regime. It shifts from a reactive, risk-averse manager in volatile downturns to a stable, coordinated orchestrator during periods of growth, validating MARS's ability to adaptively balance risk and return.

\section{Conclusion}

In this paper, we proposed MARS, a novel meta-controlled risk-aware reinforcement learning framework for portfolio management. Its core innovation lies in a two-tier architecture comprising a Heterogeneous Agent Ensemble (HAE), where each agent is assigned an explicit risk profile enforced by a Safety-Critic, and a high-level Meta-Adaptive Controller (MAC) that orchestrates the ensemble. This design allows MARS to leverage behavioral diversity to navigate changing market conditions.

Comprehensive experiments on the DJI and HSI indices demonstrate the efficacy of MARS. The framework consistently delivered higher risk-adjusted returns than established DRL baselines, and most notably, it exhibited exceptional capital preservation during the 2022 bear market by significantly minimizing drawdowns and volatility. Ablation studies confirmed that both the MAC and the heterogeneity of the agent ensemble are critical to the framework's success. These results validate that MARS provides a robust and effective solution for risk-aware portfolio management.


% ...


\section{Acknowledgments}

This work is supported by U.S. Department of Energy (Award No. DE-SC0024424) and National Science Foundation (Grant No. CNS-2340171).

\bibliography{MARS_CameraReady_final_11132025/references}

\end{document}